\documentclass[11pt,a4paper]{article}
\usepackage[final]{neurips_2022}
\usepackage{times}
\usepackage{soul}
\usepackage{url}
\usepackage{hyperref}
\hypersetup{hidelinks}
\usepackage[utf8]{inputenc}
\usepackage{caption}
\usepackage{graphicx}
\usepackage{amsmath}
\usepackage{amsthm}
\usepackage{booktabs}
\usepackage{algorithm}
\usepackage{algorithmic}
\usepackage{wrapfig}
\usepackage{multirow}
\usepackage{subcaption}
\usepackage{xcolor}
\usepackage{bbding}
\usepackage{pifont}
\usepackage{wasysym}
\usepackage{amssymb}
\usepackage[toc,page]{appendix}
\usepackage[numbers]{natbib}
\newenvironment{packed_itemize}{
	\begin{itemize}
		\setlength{\itemsep}{0pt}
		\setlength{\parskip}{0pt}
		\setlength{\parsep}{0pt}
	}{\end{itemize}}
\title{Domain-Adaptive Text Classification with \\Structured Knowledge from Unlabeled Data}

\author{Tian Li\thanks{Equal contribution.} \\
  Peking University \\
  \texttt{davidli@pku.edu.cn} \\\And
  Xiang Chen$^*$ \\
  Peking University \\
  \texttt{caspar@pku.edu.cn} \\ \And
  Zhen Dong \\
  University of California, Berkeley \\
  \texttt{zhendong@berkeley.edu} \\ \And
  Weijiang Yu \\
  Sun Yat-sen University \\
  \texttt{weijiangyu8@gmail.com} \And
  Yijun Yan \\ 
  University of California, Berkeley \\
  \texttt{bunnyyan@berkeley.edu}  \\\And
  Kurt Keutzer \\
  University of California, Berkeley \\
  \texttt{keutzer@berkeley.edu}  \\\And
  Shanghang Zhang\thanks{Corresponding author. (E-mail: {\tt\small shanghang@pku.edu.cn})} \\ 
  Peking University \\
  \texttt{shanghang@pku.edu.cn}
  }

\begin{document}

\maketitle

\begin{abstract}
Domain adaptive text classification is a challenging problem for the large-scale pretrained language models because they often require expensive additional labeled data to adapt to new domains. Existing works usually fails to leverage the implicit relationships among words across domains. In this paper, we propose a novel method, called Domain Adaptation with Structured Knowledge (DASK), to enhance domain adaptation by exploiting word-level semantic relationships. DASK first builds a knowledge graph to capture the relationship between pivot terms (domain-independent words) and non-pivot terms in the target domain. Then during training, DASK injects pivot-related knowledge graph information into source domain texts. For the downstream task, these knowledge-injected texts are fed into a BERT variant capable of processing knowledge-injected textual data. Thanks to the knowledge injection, our model learns domain-invariant features for non-pivots according to their relationships with pivots. DASK ensures the pivots to have domain-invariant behaviors by dynamically inferring via the polarity scores of candidate pivots during training with pseudo-labels. We validate DASK on a wide range of cross-domain sentiment classification tasks and observe up to 2.9\% absolute performance improvement over baselines for 20 different domain pairs. Code will be made available at \url{https://github.com/hikaru-nara/DASK}.
\end{abstract}

\section{Introduction}
\label{sec:introduction}
\begin{figure}
\centering
\includegraphics[trim=0 0 0 0, clip, width=0.48\textwidth]{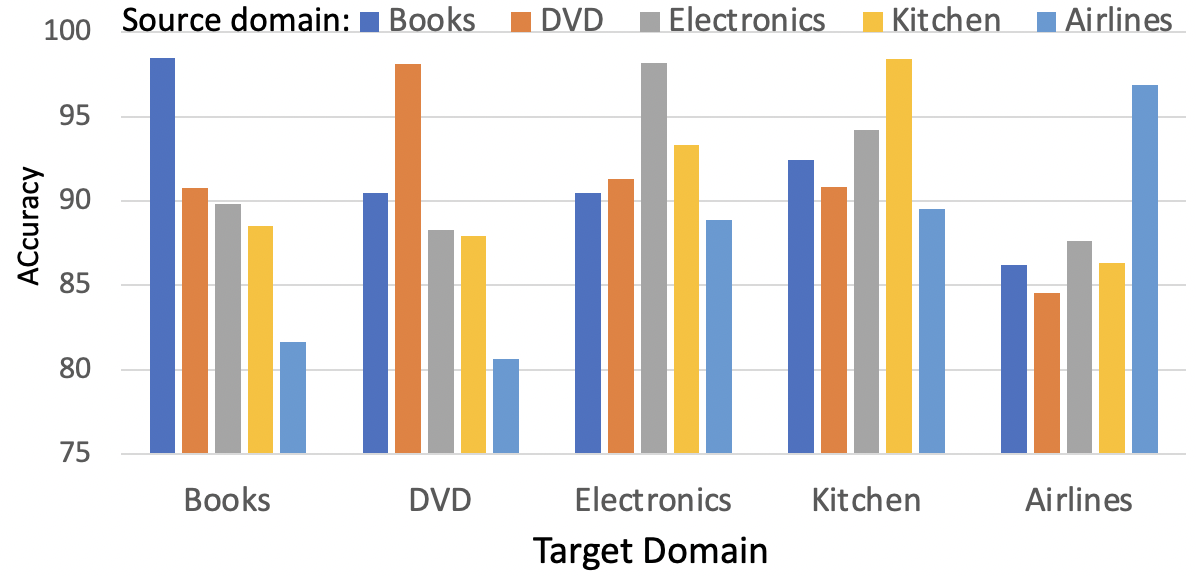}
\caption{Language models perform worse when domain shift is present. The figure shows the cross-validation results of BERT baseline models trained on 5 domains. The prediction accuracy of the models tested on the trained domain is 5-15\% higher than those tested on distant domains.}
\label{fig:baselines}
\vspace{-4mm}
\end{figure} 
Domain shift is common in many natural language processing (NLP) applications. For example, the word ``rechargeable'' is much more common in electronics product reviews than in book reviews, while the word ``readable'' is much more common in book reviews. 
Existing language models~\cite{devlin2018bert,liu2019roberta} have exhibited outstanding performance in text classification tasks, but they fail to generalize to new domains without \textit{expensive labeling and retraining} (Figure \ref{fig:baselines}). 
To break out the data constraint, some methods with unlabeled data have been proposed as follows.

Existing unsupervised domain adaptation methods for text classification can be grouped into two categories: task-agnostic methods~\cite{Long2015Learning,ganin2016domain,li2020cross,glorot2011domain} and pivot-based methods. Task-agnostic methods generally ignore the correlation among words across domains, which can contain rich semantic information in an NLP context. In contrast, pivot-based methods use domain-independent words (pivots) to bridge the domain gap by leveraging the correlations between pivots and non-pivots to learn domain-invariant features. Therefore, we would like to marry pivot-based method and pretrained language models to adapt them to novel domains. 

The most prominent pivot-based methods are Structure Correspondence Learning (SCL) and its variants~\cite{blitzer2006domain,yu2016learning,ziser-reichart-2018-pivot}. In SCL, the pivots are defined as the words that occur frequently on both source and target domains and behave in similar ways that are discriminable for the classification task\footnote{In the existing literature, the definition of pivots often uses the term ``behavior''. To put it simply, the pivots should be highly correlated to one of the labels across domains.}. The model can effectively learn domain-invariant features for pivots, but it is more challenging for the non-pivots as they have domain-specific meanings. Therefore a self-supervised auxiliary task is applied to predict the pivots from the non-pivots. As a result, SCL implicitly captures the relationships between words by recognizing co-occurrence patterns between pivots and non-pivots and uses these relationships to infer domain-invariant features for the non-pivots. 

\begin{figure*}[ht]
    \centering
    \begin{subfigure}[b]{0.33\textwidth}
        \centering
        
         \includegraphics[width=0.9\textwidth]{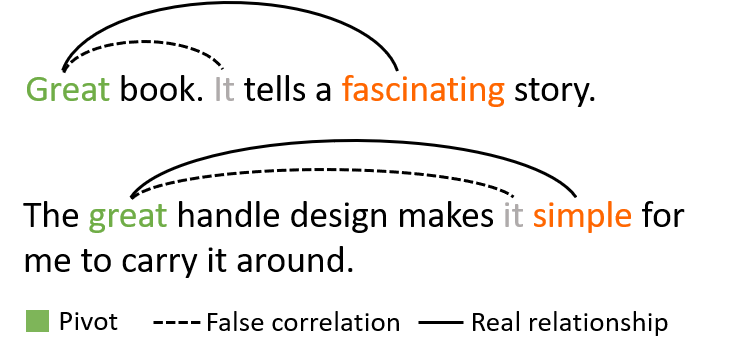}
         \caption{Structure Correspondence Learning. }
         \label{fig:SCL}
    \end{subfigure}
    \hspace{1mm}
    \begin{subfigure}[b]{0.62\textwidth}
         \centering
         \includegraphics[width=0.9\textwidth]{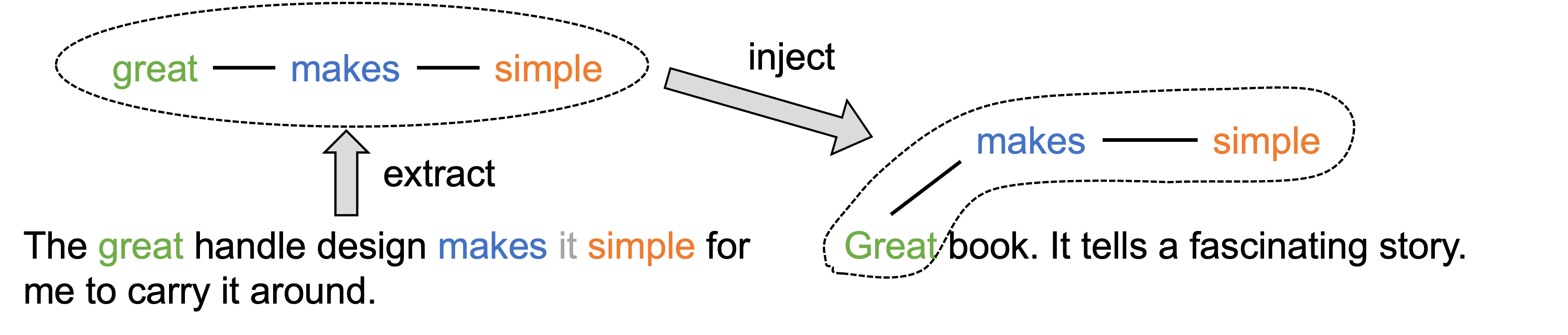}
         \caption{Two-step approach of DASK: extract and inject}
         \label{fig:KG}
    \end{subfigure}
    \caption{
    a) The example shows a pair of texts from the source domain (top) and target domain (bottom) respectively.
    Due to the frequency of ``it'' co-occurring with ``great'', the model tends to capture this false correlation. b) In contrast, DASK extracts a fact, represented by a triplet (great, make, simple), from the target domain text to filter false correlations. We inject the target domain fact into the source domain text.} 
\end{figure*}

However, SCL is limited in that it uses all non-pivots to predict the pivot terms, which leads to a noisy inference problem as very few non-pivots have a real relationship with the pivots. As a result, false correlations often occur for frequently used words such as pronouns (see Figure \ref{fig:SCL}).
Alternatively, Knowledge Graphs (KG) is an effective way to represent complex relationships between concepts in a structured format and do not solely rely on noisy co-occurrence information. Therefore, in this paper, we present a pivot-based domain adaptation method from the KG perspective. Our method, called Domain Adaptation with Structured Knowledge (DASK), follows a 2-step approach as illustrated in Figure \ref{fig:KG}. In contrast to SCL, DASK filters out false correlations by building a knowledge graph to explicitly capture the relationships between pivots and non-pivots on the target domain. 
Then during training, DASK employs a novel knowledge injection (KI) mechanism for the model to learn domain-invariant features of the non-pivots. 

Another critical drawback of SCL is that the pivots are pre-defined only on labeled source domain texts and \emph{unlabeled} target domain texts. There is little to ensure that the pre-defined pivots actually have consistent behavior across domains. To tackle this problem, DASK dynamically learns the pivots during training with pseudo-labels. A memory bank is maintained to keep track of the polarity scores of the candidate pivots on the source domain and the target domain, respectively. The words that have consistently high scores in the memory banks are collected as the pivots at the beginning of each training epoch. As we show, DASK using learned pivots outperforms the static, pre-defined pivots. 
To summarize, our major contributions are as follows:
\begin{packed_itemize}
    \item We propose DASK for text classification, 
    which injects knowledge graph facts to better leverage the relationship between words for adaptation across domains.
    \item We construct a novel knowledge graph with attention scores from BERT to explicitly capture the relationship between pivots and non-pivots on the target domain.
    \item We design the pivot-induced cross-domain knowledge injection mechanism to learn domain-invariant features for non-pivots using their relationship with pivots.
    \item We propose to maintain memory banks to learn domain-invariant pivots.
    \item We evaluate our method on the task of cross-domain sentiment classification for 20 domain pairs, where our method outperforms strong baselines by up to 2.9\%.
\end{packed_itemize}

\section{Related Work}
\label{Rel}
\subsection{Cross-Domain Text Classification}
Cross-domain text classification~\cite{glorot2011domain,ganin2016domain,ziser-reichart-2018-pivot,du2020adversarial} is a fundamental problem in domain adaptation for NLP. Before applying domain adaptation methods to complex problems, like machine translation and question answering, it was extensively researched as a simpler problem. Among text classification tasks, sentiment classification is most representative and established in terms of domain adaptation because there are standard datasets and evaluation protocols~\cite{blitzer-etal-2007-biographies,ziser-reichart-2018-pivot}. Therefore, we use sentiment classification to evaluate our method. It is important to note that our method can be easily generalized to other tasks such as topic labeling, news classification and so forth.

Methods for cross-domain text classification can be roughly categorized into two classes: task-agnostic methods, and pivot-based methods. 
The former includes divergence minimization~\cite{Long2015Learning,sun2016return,he2018adaptive}, stacked denoising auto-encoders~\cite{glorot2011domain}, discriminative adversarial training~\cite{ganin2016domain,tzeng2017adversarial}, instance reweighting~\cite{chen2021wind} and so forth. There are also some works that combine task-agnostic methods with NLP-specific approaches or models~\cite{ghosal2020kingdom,du2020adversarial}. 
In contrast, pivot-based methods are different from them in two aspects: 1) they make use of the relationship between pivots and non-pivots to help learn discriminative feature, 2) they focus on learning word-level features instead of instance-level features, which align with the nature of language. These two points make pivot-based methods stand out on NLP tasks.

\subsection{Knowledge Graph}
\paragraph{Knowledge Graph Construction.}
KG can be constructed in a supervised manner, such as DBPedia \cite{lehmann2015dbpedia} and Wikidata \cite{vrandevcic2014wikidata}; semi-supervised manner such as Google's Knowledge Vault \cite{dong2014knowledge}; or unsupervised manner, such as MAMA \cite{wang2020language}. 
Among them, we are most interested in MAMA because it uses learned knowledge stored in pre-trained language models without human supervision to construct a KG. With a forward pass of BERT, it collects the words or phrases in the input that have high attention scores with each other to form a candidate fact, which is represented by a triplet \emph{(head, relation, tail)}. 
Hence it is an easy fit for our knowledge-injected language model P-BERT. We can initialize P-BERT with the same weights from BERT used in MAMA, so that it becomes easier for P-BERT to interpret the knowledge graph facts that are constructed by MAMA.

\paragraph{Knowledge-Injected Transformers.} 

Since the emergence of pre-trained language models like BERT~\cite{devlin2018bert,liu2019roberta}, many works have followed up to incorporate external knowledge into them~\cite{wang2020k,zhang2019ernie,wang2021kepler,liu2020k}. Typically, these works either fuse textual features and factual features at the end of the corresponding encoders to enhance language understanding~\cite{wang2021kepler,zhang2019ernie}, or use adapters for injection of multiple knowledge sources with the original model fixed~\cite{wang2020k}. Although most of them tried to utilize open-domain knowledge graphs or knowledge bases, K-BERT~\cite{liu2020k} takes domain-specific knowledge as input along with text, and thus is particularly suitable for our task of domain adaptation. Also, by using a flattened sentence tree and the visible matrix, it is compatible to BERT and free from additional pre-training. Therefore, to promote domain transfer with pivots, we develop K-BERT into P-BERT by replacing its KI module with our Pivot-induced Cross-domain Knowledge Injection. 

Beyond knowledge-injected transformers, another line of related works is tree-based transformers for processing tree-structured data in general~\cite{Shiv2019NovelPE,sun2020treegen}. Despite their power in learning from arbitrary tree-structured data, they are not directly compatible to BERT.

\begin{figure*}[t]
    \centering
    \includegraphics[width=0.99\textwidth]{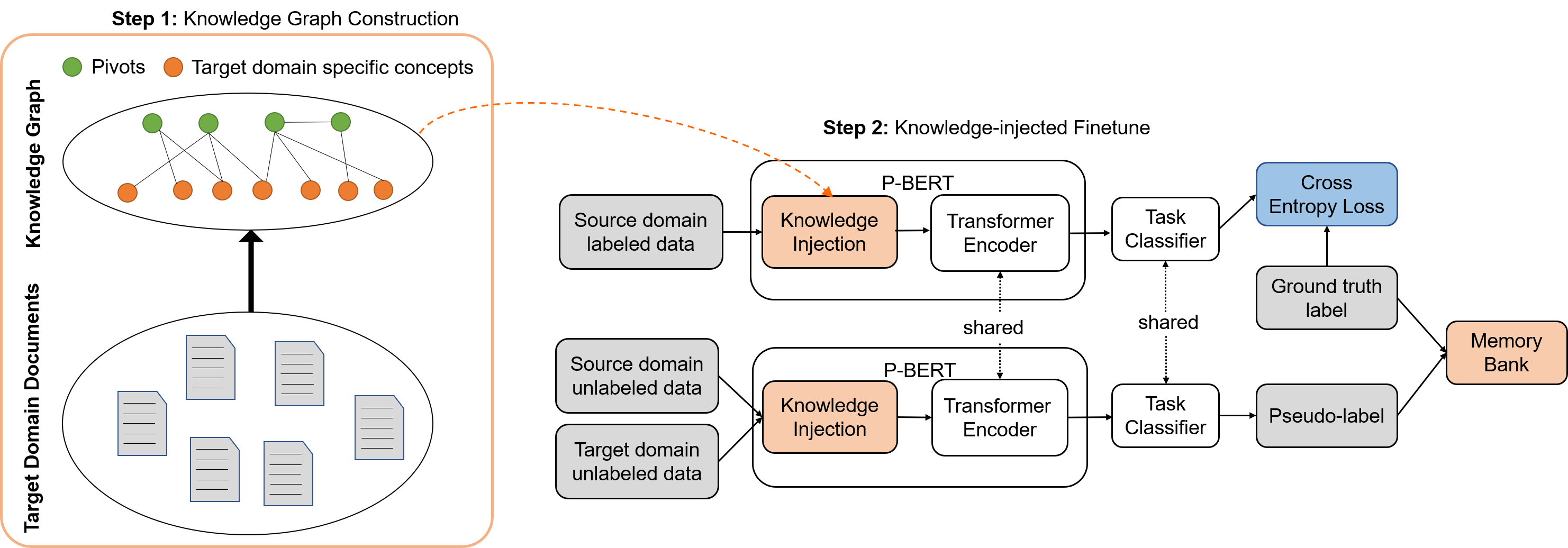}
    \caption{Illustration of DASK. DASK consists of two steps. In step 1 we construct a knowledge graph from target domain unlabeled data. In step 2 we finetune the model on knowledge-injected data and learn the pivots with memory bank.}
    \label{fig:pipeline}
\end{figure*}

\section{Method}
\label{sec:method}

DASK aims to facilitate domain adaptation with structured knowledge from the target domain. To structure the knowledge in the target domain, we build a knowledge graph to capture the relationship among pivots and non-pivots and inject them into source domain data. This is helpful because the model is able to learn domain-invariant feature for non-pivots (domain-specific) by inferring from their relationship with domain-shared pivots. Specifically, DASK follows a 2-step approach as follows (see Figure \ref{fig:pipeline}):
\paragraph{Step 1.}
We construct a knowledge graph (KG) from the target domain texts to model the relationship between pivots and non-pivots. Specifically, it involves 1) extracting candidate facts from sentences, 2) filtering low-confidence facts. Section \ref{sec:graph construction} shows details of how we construct the KG. 
\paragraph{Step 2.}
After constructing the knowledge graph, we start knowledge-injected finetuning on the source and target domains jointly. At each training step, we obtain labeled texts from the source domain, unlabeled texts from the source domain and target domain. The inputs are fed into P-BERT where all of them are injected with pivot-relevant facts from the knowledge graph (section \ref{sec:knowledge injection}), and then forwarded to the transformer encoder and the linear classifier sequentially. Finally, the predictions of labeled data are used to compute the cross-entropy loss with the ground truth labels. Meanwhile, the predictions of unlabeled data are leveraged to generate pseudo-labels for pivot learning (section \ref{sec:pivot learning}). The pseudo-labels of unlabeled data and ground truth labels of labeled data give updates to the polarity scores in the memory banks. 
Note that during inference the memory bank is not updated. 

\subsection{Knowledge Graph Construction}
\label{sec:graph construction}
Prior to training, we build a KG on the target domain corpus to capture the relationship between pivots and non-pivots.
\paragraph{Candidate Fact Extraction.}
We extract candidate facts on a sentence level. An input text is decomposed to a list of sentences. We feed each sentence into BERT to get the attention matrix in the last transformer encoder layer. As a pre-processing step, the multi-head attention is averaged into single-head so that one pair of words only corresponds to one scalar attention value (if a word consists of multiple tokens, the attentions of the tokens are also averaged). Denote the pre-processed attention matrix as $M$.
For each pivot $p$ in the sentence, we search for the words $w_1, w_2$ that have the highest and second highest attention with $p$. Then the fact is formed as the triplet of $w_1, w_2, p$ in their \textit{original order} in the sentence. In addition, each fact is assigned a confidence score $M[p][w_1]+M[p][w_2]$.

\paragraph{Filtering.}
The candidate facts are filtered according to their confidence scores. Those whose confidence scores are under a threshold are removed from the knowledge graph. 

Note that although we learn pivots dynamically during training, those pivots all come from a large pivot pool that is constructed before training (see section \ref{sec:pivot learning}). We construct the knowledge graph according to the pivot pool so that we do not have to update the knowledge graph as we learn new pivots and eliminate old ones. In the appendix, we showcase some qualitative results of a KG that we constructed in this way and compare it with the ConceptNet subgraph. 

\subsection{Knowledge Injection}
\label{sec:knowledge injection}
In order to utilize the rich semantics in the KG to learn domain-invariant features for the non-pivots, we propose Pivot-induced Cross-domain Knowledge Injection (PCKI) to inject knowledge facts into input texts. On the other hand, we would like to use a pretrained language model as the feature extractor, so we borrow ideas from K-BERT~\cite{liu2020k} to let the language model understand the structure of knowledge injected input. As a whole, our model P-BERT is composed of the knowledge injection module and the subsequent transformer encoder, as illustrated in Figure \ref{fig:pipeline}.

In the knowledge injection module, for each pivot in the input text, we search for the facts relevant to it and inject them into the text, forming a tree structure. Figure \ref{fig:KG} shows an example for a knowledge injected text. In the example, ``Great'' is a pivot in the sentence on the right, and we have the fact triplet ``(great, makes, simple)'' in the KG that we extract from the sentence on the left. The triplet is then appended to ``great'' as a branch. By injecting facts extracted from the target domain into labeled source domain texts, we embed the target domain non-pivots in a labeled context, conditioning on the related pivots. Therefore, our transformer encoder is able to infer their features according to their relationships with pivots, under the supervision of source domain labels. 

\begin{figure*}[t]
    \centering
    \includegraphics[width=0.99\textwidth]{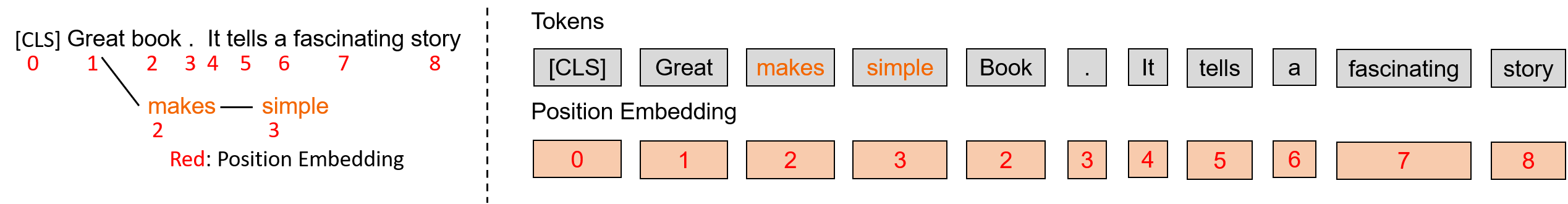}
    \caption{ 
    Continuing the example in Figure~\protect\ref{fig:KG}, we inject the fact triplet \emph{(great, makes, simple)} to the main sentence forming a tree structure (left). On the right, we flatten the tree into a sequence and use the depth of the tokens in the tree as their position embedding index. The highlighted words in {\color{orange}orange} are the injected fact.
    }
    \label{fig:position embedding}
\end{figure*}

To feed the knowledge-injected text to the transformer encoder while keeping the structure information, we flatten it into a token sequence, and use position embedding to recover its structure~\cite{liu2020k}. Specifically, as illustrated in Figure \ref{fig:position embedding}, the injected words are inserted before or after the corresponding pivot following their order in the triplet so as to get the flattened token sequence. Meanwhile, to recover the tree structure, we assign the index for position embedding of each token as its depth in the tree. In the example, ``makes'' and ``book'' are assigned the same position embedding index 2.
In this way, the token sequence is fed into the transformers, with the position embedding informing its structure. Besides, a visible matrix is applied to the self-attention modules in the transformers to control knowledge noise. We refer readers to K-BERT~\cite{liu2020k} for more details on visible matrix. 

\subsection{Pivot Learning}
\label{sec:pivot learning}
As a pivot-based method, DASK heavily relies on the domain-shared pivots. In this section, we describe how we select and learn the pivots with memory banks.

Recall that pivots are defined as the words that behave similarly in the source and target domains. Following previous works, we represent the behavior of a word with the labels of the texts in which it appears. The principle is that the label distribution of a pivot should be \textit{low-entropy} (biased towards one label), and \textit{consistent} across domains. While it is applicable to most classification tasks, for the binary sentiment classification task we focus on, we define a polarity score $p(w,D)$ to measure more easily how much the label distribution of $w$ is biased towards the labels on domain $D$: 
\begin{equation}
    p(w,D)=\frac{\left|\{l=1|l\in b(w,D)\}\right| - \left|\{l=-1|l\in b(w,D)\}\right|}{|b(w,D)|}
\label{equation:2}
\end{equation}
where $b(w,D)$ is the label set of $w$ on $D$. And thus the behaviour on the domain pair $(S,T)$ can be characterized as:
\begin{equation}
    \bar p(w, (S,T)) = \frac{|p(w,S)+p(w,T)|}{2}
    \label{equation:3}
\end{equation}
Taking the absolute average ensures that a word of high score must be biased towards the same label on both domains.

Prior to training, we collect a large pool of candidate pivots from the vocabulary which contains words that appear frequently on both domains. Other words are not taken into account because the precondition is that pivots should be frequent on both domains.
To track the polarity scores of the candidate pivots, we build memory banks for the source domain and the target domain respectively. 
Both of them are initialized by scores computed with equation \ref{equation:2} constrained on source domain labeled data. Then we compute the absolute mean polarity score of the candidate pivots following equation \ref{equation:3} and take the $K$ candidate pivots with the highest score as the initial pivots.

During training, at each training step, we acquire pseudo-labels for the unlabeled texts if the prediction confidence in the softmax logits is over a threshold. The pseudo-labels of the unlabeled source domain inputs, and the ground truth labels of the labeled source domain inputs are used to update the source memory bank, while the pseudo-labels of the target domain inputs are used to update the target memory bank. The update is carried out in a temporal difference style: For each candidate pivot, if it is in the text from domain $D$ labeled as $l\in \{1,-1\}$, then 
\begin{equation}
    p(w,D)\leftarrow \alpha \cdot p(w,D) + (1-\alpha) \cdot l
\end{equation}
where $\alpha$ is the update rate.
In this way, we manage to estimate the behavior of words more accurately on the domains, so that domain-invariant pivots are acquired. At the beginning of each epoch, we compute the absolute mean scores of the candidate pivots following equation \ref{equation:3} and take the top-$K$ candidate pivots as learnt pivots. The pivots are kept fixed in the middle of an epoch to avoid overly frequent pivot changes.

\section{Experiments}

To evaluate the efficacy of our proposed method, we extensively experiment on two cross-domain sentiment classification datasets. We compare our method with multiple baselines on 20 domain pairs and provide justification for the improvement of our method. Moreover, we carry out ablation study to respectively assess the effect of knowledge injection and dynamic pivot learning. 

\subsection{Experiment Settings}
\label{sec:experiments settings}

\paragraph{Datasets.}
We use the standard Amazon-product-review\footnote{Dataset can be found at \url{http://www.cs.jhu.edu/~mdredze/datasets/sentiment/index2.html}} dataset~\cite{blitzer-etal-2007-biographies}. It contains four types of product reviews: books(B), dvd(D), electronics(E), kitchen(K), which form the four domains in the dataset. Besides, following PBLM~\cite{ziser-reichart-2018-pivot}, we also introduce the Airlines dataset\footnote{Dataset and process procedures can be found at \url{https://github.com/quankiquanki/skytrax-reviews-dataset}}.

\begin{table}[t]
    \centering
    \begin{tabular}{c |c c c c| c c}
    \toprule[1.5pt]
         \multirow{2}{*}{S$\rightarrow$ T} & \multicolumn{6}{c}{BERT}\\
         \cline{2-7} & Base & HATN & DANN & DAAT  & DASK & DASK+SCL\\
         \midrule[1pt]
         B$\to$ E & 90.50 & 87.21 & 91.67 & 89.57 & 91.95 & \textbf{92.30}  \\
         B$\to$ D & 90.45 & 89.36 & 89.93 & 89.70 & 90.55 & \textbf{90.90}  \\
         B$\to$ K & 92.46 & 89.41 & 92.80 & 90.75 & \textbf{92.85} & 92.75  \\
         E$\to$ B & 89.85 & 87.10 & 89.19 & 88.91 & {89.70} & \textbf{90.00} \\
         E$\to$ D & 88.30 & 88.81 & 88.49 & \textbf{90.13} & 88.65 & 89.20 \\
         E$\to$ K & 94.20 & 92.01 & 94.54 & 93.18 & 94.35 & \textbf{94.65} \\ 
         D$\to$ B & 90.75 & 89.81 & 91.37 & 90.86 & 91.20 & \textbf{91.85} \\ 
         D$\to$ E & 91.30 & 86.99 & 91.52 & 89.30 & 88.70 & \textbf{92.40} \\
         D$\to$ K & 90.85 & 87.59 & 92.16 & 90.50 & 91.80 & \textbf{92.35} \\
         K$\to$ B & 88.50 & 89.36 & 89.38 & 87.98 & \textbf{90.15} & {89.75} \\
         K$\to$ E & 93.34 & 90.31 & 93.15 & 91.72 & 92.80 & \textbf{93.35} \\
         K$\to$ D & 87.90 & 87.89 & 88.89 & 88.81 & 88.40 & \textbf{89.45} \\
         \emph{Average} & 90.70 & 88.69 & 91.09 & 90.12 & 90.92 & \textbf{91.59}   
         \\   
    \bottomrule[1.5pt]
    \end{tabular}
    \setlength{\abovecaptionskip}{5pt}    
    \caption{Cross-domain sentiment classification accuracy on 12 domain pairs from Amazon-product-review dataset. Our method is able to outperform all the strong baselines on all domain pairs with the only exception of E$\to$D. For BERT-HATN and BERT-DAAT we use numbers reported by \protect\cite{du2020adversarial}.}
    \label{tab:major_results}
\end{table} 

\begin{table*}[t]
    \begin{subtable}[h]{0.35\textwidth}
        \centering
        \resizebox{0.98\textwidth}{!}{
        \begin{tabular}{c|c c |c c}
        \toprule[1.5pt]
        \multirow{2}{*}{S$\rightarrow$ T} &  \multicolumn{4}{c}{BERT}\\
        \cline{2-5}
             & Base & DANN & DASK & DASK+SCL \\
        \hline
             A$\to$ B & 81.65 & 81.50 & 82.10 & \textbf{84.15} \\
             A$\to$ E & 88.85 & \textbf{89.53} & 89.35 & 89.10 \\
             A$\to$ D & 80.60 & 82.74 & \textbf{83.15} & 82.85 \\
             A$\to$ K & 89.50 & 89.53 & 89.90 & \textbf{90.00} \\
             B$\to$ A & 86.18 & 86.66 & 86.30 & \textbf{86.70} \\ 
             E$\to$ A & 87.60 & \textbf{87.90} & 87.30 & \textbf{87.90} \\
             D$\to$ A & 84.55 & 86.71 & 84.85 & \textbf{86.75} \\
             K$\to$ A & 86.30 & 86.56 & 86.50  & \textbf{86.80} \\
        \emph{Average}& 85.65 & 86.39 & 86.12 & \textbf{86.78} \\
        \bottomrule[1.5pt]
        \end{tabular}
        }
        \caption{
        }
        \label{tab:airlines}
    \end{subtable}
    \hfill
    \begin{subtable}[h]{0.6\textwidth}
        \centering
        \resizebox{0.98\textwidth}{!}{
        \begin{tabular}{c|c|c|c c c c c} 
        \toprule[1.5pt]
        \multirow{2}{*}{\textbf{PCKI}} & \multirow{2}{*}{\textbf{SCL}} & \multirow{2}{*}{\textbf{Dynamic}} & \multicolumn{5}{c}{\textbf{Accuracy}} \\
        \cline{4-8} & & & A$\to$ B & B$\to$E & E$\to$K & K$\to$D & D$\to$A \\
        \hline
         & & & 81.65 & 90.50 & 94.20 & 87.90 & 84.55 \\
        &  \checkmark &  & 81.85 & 90.95 & 94.40 & 87.95 & 84.90\\
        \checkmark&   &  & 82.10 & 91.90 & 94.25 & 88.20 & 84.75 \\
        \checkmark & \checkmark  &   & 82.95 & 92.05 & 94.60 & 88.40 & 86.05\\ 
        \checkmark &  \checkmark  & \checkmark & \textbf{84.15} & \textbf{92.30} & \textbf{94.65} & \textbf{89.45} & \textbf{86.75}\\
        \bottomrule[1.5pt]
        \end{tabular}
        }
    \setlength{\abovecaptionskip}{5pt}
    \caption{}
    \label{tab:ablation}
    \end{subtable}
    \caption{a) Cross-domain sentiment classification accuracy on the 8 domain pairs between airlines dataset and 4 domains from Amazon-product-review dataset. b) Ablation study on PCKI, SCL and dynamic memory bank. We did experiments on 5 domain pairs.}
    \vspace{-2mm}
\end{table*}

We evaluate our method on all 20 domain pairs involving the five domains. Each time before training, we randomly sample 400 labeled source domain data for dev set, and the rest 1600 along with all unlabeled data from source and target domain are used for training. During testing on the 2000 target labeled data, we stop the memory bank update, and use the pivots learnt at the best training epoch. 

\paragraph{Baselines.}
We use BERT~\cite{devlin2018bert} as the feature extractor. Apart from directly finetuning BERT on the source domain, we compare our method with multiple strong baselines incorporating methods proposed by previous works. The baselines are listed as follows:
\begin{packed_itemize}
    \item \textbf{Base}: BERT trained on the source labeled data and directly test on the target labeled data.
    \item \textbf{HATN}: BERT combined with HATN proposed by \cite{li2018hierarchical}. It is a prominent variant of Structure Correspondence Learning (SCL).
    \item \textbf{DANN}: BERT combined with the popular adversarial training method, DANN~\cite{ganin2016domain}. 
    \item \textbf{DAAT}: BERT-DAAT proposed by~\cite{du2020adversarial}. It combines target domain post-training along with domain adversarial training to boost domain adaptation.
\end{packed_itemize}

To show the compatibility of DASK with existing methods, we stack SCL on top of our method. Since there was no existing work applying SCL on transformers, we manufactured a training scheme for SCL that mimics MLM in BERT~\cite{devlin2018bert} pretraining, that is, to replace the pivots as [\texttt{MASK}] tokens and ask the model to recover the masked tokens given the context.

\paragraph{Hyperparameter Tuning.}
For all the methods in our experiments, we set the learning rate to 2e-5, warmup $0.1$, batch size 32, and select weight decay in $\{\text{1e-4, 2e-4, 3e-4}\}$. For adversarial training, we select $\gamma$ in the gradient reversal layer from $\{0.15,0.25,0.5,0.75,1.0\}$. For KG fact filtering, we set the confidence threshold in the interval $[0.1,0.45]$, according to the confidence distribution. Besides, we leverage the stopwords in the NLTK library and discard the facts that include the stopwords. For SCL, we apply a balance factor $\lambda$ to the pivot-prediction loss, and choose its value from $[0.1,0.5]$. To make the learning progress smoother, we also use a linear warmup to $\lambda$ and set the warmup rate to $0.1$. In addition, since our sentiment classifier is trained on the source labeled data, and the size of labeled data is 6-20 times smaller than that of unlabeled data, our model see each labeled data for 6-20 times in one epoch. In order to prevent overfitting, we update the sentiment classifier once in 5-11 training steps. For the dynamic memory bank, we update the pivots every $10$ training steps, and the top 500 words are selected as pivots.The learning rate of word sentiment score is set to 1e-4 or 2e-4, and the pseudo-labeling confidence threshold to $0.9$.

\subsection{Experimental Results}
Table \ref{tab:major_results} shows the performance of our method on the 12 domain pairs from Amazon-product-review dataset, compared with multiple baselines.

Among all those methods, we observe that DASK+SCL gives the best performance on average. It outperforms the Base model by 0.89\%, and beats the other strong baselines by 0.50\%-2.90\% on an average basis over all domain pairs. On the other hand, DASK also has competitive performance. It is better than DAAT and HATN by 0.8\% and 1.43\%, outperforms the Base Model by 0.22\%. Although its average accuracy is slightly lower than DANN, it gives the best performance on the B$\to$ K and K$\to$B settings. 

Moreover, we conduct experiments on 8 domain pairs involving the Airlines dataset. Table \ref{tab:airlines} shows the performance of our method on those settings. Although this setting is more difficult than domain pairs within Amazon-product-review dataset, we are still able to outperform Base on all of the domain pairs, by 1.13\% on average. This shows that, by exploiting correlation among words, our method can help the model capture relationship between pivots and non-pivots even from distant domains. Besides, our method betters DANN by 0.39\%. Although this is not a thick margin, it is non-trivial because 1) DANN itself is a very competitive method, 2) we do not expect to beat DANN by any larger margin, but to propose a strong baseline for domain adaption from a KG perspective, 3) our method is orthogonal to DANN and thus can be coupled with DANN to achieve even better performance.

Note that our Base model performs better than the previous state-of-the-art BERT-DAAT. This owes to hyperparameter tuning where BERT-DAAT uses 1e-2 for weight decay while we use 1e-4. It is not open-sourced so we cannot reproduce their results for better comparison. Nevertheless, it does not mean our performance gain comes from hyperparameter tuning as we analyze in the ablation studies below.

\begin{table}[ht]
    \centering
    \begin{tabular}{c|c|c} 
    \toprule[1.5pt]
    \textbf{KI method} & \textbf{KG} &  \textbf{Accuracy} \\
    \midrule[1pt]
    
    normal & subgraph &  80.45 \\
    normal & learnt &   77.45 \\
    PCKI & subgraph  &   79.50 \\
    
    PCKI & learnt  &   \textbf{82.10} \\ 
    \midrule[1pt]
    \multicolumn{2}{c|}{Base} & 81.65 \\
    \bottomrule[1.5pt]
    \end{tabular}
\setlength{\abovecaptionskip}{5pt}
\caption{Ablation studies on knowledge injection mechanism and KG construction method. All experiments are done on the A$\to$B domain pair.}
\label{tab:ablation2}
\end{table}

\paragraph{Ablation Study.}
As a sanity check, we performed experiments to study the effect of our choices of KI method and the knowledge graph. We compare PCKI versus normal KI method which inserts knowledge to every word possible, and our learnt KG versus the ConceptNet subgraph. From table~\ref{tab:ablation2}, it can be observe that none of the configurations works other than PCKI plus learnt KG, our proposed method.

More importantly, to analyze how much each part of our method contribute to the performance gain, we conduct ablation study on three aspects: PCKI, SCL and pivot learning (Dynamic). We did experiments on 5 domain pairs as shown in table \ref{tab:ablation}. From the results, we have the following observations: 1) PCKI and SCL are both able to help DA independently, and PCKI generally works better than SCL, 2) Jointly applying them can further boost the performance, which implies that they complement each other in the ability of exploiting the correlation among pivots and non-pivots, 3) dynamically learning the pivots improve the performance by a large margin compared to using a set of pre-defined pivots. This supports that the learnt pivots are more beneficial to domain adaptation than the pre-defined ones. 

\subsection{Qualitative Results}
\begin{table*}[t]
    \centering
    \begin{tabular}{c|l|l}
        \toprule[1.5pt]
         \textbf{Entity} & \textbf{ConceptNet Subgraph} & \textbf{Learnt KG}  \\
         \midrule[1pt]
         \multirow{7}{*}{great}
         & (great, related to, good) & (looks, surprisingly, great) \\
         & \emph{(great, related to, alexander)} & (great, is, awesome)  \\
         & (great, similar to, large) & (great, is, excellent) \\
         & (mega, related to, great) & \textbf{(great, save, \$)} \\
         & (great, related to, awesome) & \textbf{(really simple, got, great)} \\ 
         & \emph{(rocking, related to, great)} & \textbf{(easy, seems, great)} \\
         & \emph{(lies, related to, great)}  & (also, looks, great) \\
         \hline 
         \multirow{7}{*}{simple}
         & \emph{(simple, related to, unsophisticated)} & \textbf{(simple, is, amazing)} \\
         & \emph{(five needled, similar to, simple)} & \textbf{(charging, simple, quick)}\\
         & (simpler, form of, simple) & \textbf{(excellent condition, putting, simple)} \\
         & \emph{(simple, synonym, unsuspecting)} & (plain, and, simple) \\
         & \emph{(cakewalk, related to, simple)} & (the setup, fairly, simple) \\
         & (easy, related to, simple) & \textbf{(amazon, simple, fast)} \\
         & (plain, related to, simple) & (makes, it, simple)\\
        \bottomrule[1.5pt]
    \end{tabular}
    \caption{Visualization of the triplets in learnt knowledge graph compared to Conceptnet subgraph. Both graphs are extracted for the domain pair B$\to$E. The \textbf{bold} triplets indicate a relationship between non-pivots on the E domain and pivots between B$\to$E domain pair. The \emph{italicized} triplets are knowledge noise that irrelevant to the target domain. The results show that our learnt knowledge graph better models the relationships between pivots and relevant non-pivots and avoids irrelevant knowledge noise.}
    \label{tab:graph quality}
\end{table*}
\subsubsection{Knowledge Graph}
To demonstrate the advantage of the learnt KG over the ConceptNet subgraph in modeling relationship between pivots and non-pivots for domain adaptation, we visualize an example of ConceptNet subgraph and our KG in Table \ref{tab:graph quality}. Each cell contains the triplets involving the concept on the left. For example, the top-left grid contains the triplets in ConceptNet subgraph that involves ``great'', etc. In the table, the ConceptNet subgraph is the one-hop subgraph induced by words on the electronics (E) domain corpus and the learnt KG is the result of applying our method described in \ref{sec:graph construction} to the B$\to$E domain pair. Here, we consider the concepts ``great'' and ``simple'' because ``great'' is a pivot that carries positive sentiment, while ``simple'' is a positive non-pivot on E domain that carries positive sentiment (``simple'' could mean ``unsophisticated'' in book reviews, which is a negative word for stories.). We want to examine how our learnt KG explicitly models the relationship between pivots and non-pivots and therefore benefits domain adaptation with PCKI.
From table \ref{tab:graph quality}, we notice the following facts:

In both ConceptNet subgraph and our KG, ``great'' is related to other positive pivots (good, awesome, large, excellent), and some domain-specific concepts.
However, in ConceptNet subgraph, the related domain-specific concepts is not necessarily related to electronic products. This is because ConceptNet is a commonsense KG extracted from open-domain corpus such as Wikipedia. Therefore, concepts like ``alexander'', ``rocking'', and ``lies'' that are irrelevant to electronics would introduce noise when used in knowledge injection.

In contrast, our KG only contains concepts that occur in electronics domain corpus, which avoids knowledge noise. Besides, in our KG, ``great'' is usually related to other sentiment indicating words or phrases, such as ``save \$'', ``really simple'', ``easy'' and so forth. Notably, those words or phrases probably does not convey consistent sentiment on source domain (Books), and thus the model cannot learn discriminative features for them without knowledge injection. Coupled with PCKI, these relationships will enable our model to understand those target domain-specific concepts and to learn discriminative features for them.

Similarly, ConceptNet relates ``simple'' to synonyms like ``easy'' and various irrelevant domain-specific concepts such as ``cakewalk'' and ``five needled''. In contrast, our  KG relates it to pivots such as ``amazing'' and ``excellent condition''. These connections between pivots and non-pivots are able to help model learn discriminative feature with PCKI.

Moreover, our KG is more flexible and specific than ConceptNet subgraph. The vast majority of ConceptNet subgraph relations are ``related to'', ``similar to'', but the relation in our Bilevel KG has better diversity. Conventionally, the relation set of a KG is designed by human experts and is thus relatively limited. In contrast, we extract Bilevel KG with a language model from the domain corpus. Although some of the extracted relations are not verb, they actually indicate anonymous relationship with complex semantic meaning that should not be plausibly marked with ``related to''. For example, (great, save, \$) actually indicates anonymous relationship between ``great'' and ``save \$''.
Those anonymous relationships greatly enrich the content of the knowledge graph and avoids using plausible terms like ``related to'' which could induce knowledge noise.

In conclusion, the ConceptNet subgraph is noisier than the learnt KG. The anonymous relationship in the learnt KG taps the potential of an NN language model to interpret it.

\begin{table*}[t!]
    \centering
    
    \resizebox{0.95\textwidth}{!}{
    \begin{tabular}{c|c|c}
        \toprule[1.5pt]
         S$\to$T & \textbf{Initial Pivots} & \textbf{Learned Pivots} \\
         \midrule[1pt]
         \multirow{4}{*}{B$\to$E}
         & anything, \emph{pages,} \textbf{comfortable}, got, wanted, left,& \textbf{great, best, love,} history, personal, lives,    \\
         & completely, \emph{paper,} went, began, pick, seems, & involved, \textbf{larger, enjoyed, trust,} eye, also,  \\
         & \textbf{trouble, average,} add, example, stand, & \textbf{thank, definitely,} knowledge, \textbf{honest,} means, \\
         & fell, effort, self, expectations, virtually, \emph{artist} &  leader, \textbf{critical,} reviewed, draw, \textbf{sweet,} drawing  \\
         \hline 
         \multirow{5}{*}{E$\to$K}
         & back, tried, minutes, different, year, cut, & \textbf{well, works,} use, \textbf{good, nice,} used, also, \textbf{easily,} \\
         & anything, \textbf{work}, \emph{charge,} goes, service, months, & \textbf{like, recommend, quality,} hand, need, music, \\
         & several, received, products, days, four, imagine, & \textbf{heavy, beautiful,} far, sound, included, paper, \\
         & selling, point, hot, track, happen, something, whole, & including, \textbf{fairly,} taking, watch, operation, home, \\
         & went, experience, \textbf{neither,} ok, pass, half, \emph{image} & become, turning, includes, \textbf{lot,} transfer, connects\\
         \hline
         \multirow{6}{*}{K$\to$D} 
         & back, money, end, thought, left, trying, second, & \textbf{well,} one, \textbf{every, like, good,} makes, really, set, \\
         & check, coming, stand, help, alone, times, instead, & also, \textbf{recommend, especially,} seen, \textbf{ever, beautiful,} \\
         & \textbf{huge,} \emph{sheets, months,} forget, \emph{temperature,} count & \textbf{always,} use, \textbf{friend, kind,} etc, \emph{pans,} might, \emph{tea,} want\\
         & seconds, \emph{rust,} saying, plan, \emph{ingredients,} twice, & mostly, wife, food, almost, let, version, long, heat, \\
         & picture, \emph{loaf,} putting, \emph{delivery, turned, brown,} went, & much, handles, glasses, party, mix, duty, \\
         & behind, directions, \textbf{fair, correct,} elsewhere, sort & green, come\\
        \bottomrule[1.5pt]
    \end{tabular}
    }
    \caption{Qualitative results of dynamically learnt pivots at the end of training compared to the initial pivots. To be clear and concise, we do not show their intersection but only show their difference. \textbf{Bold} ones are the words we would probably deem as pivots from human instinct. \emph{Italicized} ones are the words that bias to the source domain, i.e. source domain-specific concepts. This visualization demonstrates that the pivots that we learn from dynamic memory banks are more consistent with human sense and 
   more domain-general than those pre-defined by rules in previous works.}
   
    \label{tab:learnt pivots}
\end{table*}
\subsection{Dynamically Learned Pivots}
To demonstrate the effect of pivot learning, we visualize the initial pivots and our learnt pivots on B$\to$E, E$\to$K and K$\to$D domain pairs in table~\ref{tab:learnt pivots}. The ``initial pivots'' are the pivots from the initialized memory banks before training. Note that they are identical to  the pre-defined pivots used by previous works~\cite{ziser-reichart-2018-pivot}. The ``learnt pivots'' are the pivots from the memory banks at the end of training (see section~\ref{sec:pivot learning}). We evaluate the quality of a set of pivots by two criteria. Firstly, we expect the pivots are sentiment indicating, which means that we expect the pivots to have significant sentiment polarities. Secondly, to promote the domain transfer, we aim to prevent the pivots from being biased to the source domain. We report the difference of the two sets of pivots in table ~\ref{tab:learnt pivots} \textit{without cherry picking}. From table~\ref{tab:learnt pivots}, we have the following observations:
\begin{itemize}
    \item \textbf{Bold} words are some typical pivots that we judge from human instinct. The number of this kind of words, such as ``great'', ``best'' and ``love'', is greatly increased by dynamically learning the pivots with the memory bank.
    \item \textit{Italicized} words are some source domain-specific concepts which are biased to the source domain and may degrade the performance in the target domain. We observe less source domain-specific concepts after pivot learning. This shows that our method is able to learn domain-general pivots in order to better promote the domain transfer.
\end{itemize}
These observations owe to our dynamic memory banks, which maintain the polarity scores of the common words between the two domains. When computing the initial polarity scores of the words, only labeled data on the source domain is available, which makes the induced pivots prone to bias to the source domain. In contrast, our memory bank is dynamically updated during training not only with pseudo-labels of the unlabeled text on both source and target domains, but also with the ground-truth labels of the labeled text. Therefore, our memory banks obtain comprehensive understandings of the words during training, and thus the learnt pivots will be more accurate and less biased to the source domain. 

\section{Conclusion}
In this paper, we proposed Domain-Adaptive text classification with Structured Knowledge (DASK) which elegantly marries pivot-based methods, knowledge graph and pretrained language models. It builds a knowledge graph from the target domain data to model the relationship between pivots and non-pivots. 
During training, DASK injects knowledge into the input sentence by means of pivot-induced cross-domain knowledge injection, and P-BERT is proposed to encode the knowledge-injected sentence.
In this way, DASK learns domain invariant features by capturing the relationship between pivots and target domain-specific non-pivots. 
We conducted experiments on cross-domain sentiment classification and showed that DASK outperformed strong baselines in prediction accuracy by up to 2.9\% on 20 domain pairs. 
Despite that our experiments are focused on sentiment classification, DASK generalizes to other cross-domain text classification tasks, as long as pivots are well-defined for the task. We reserve it as future work to extend DASK to other tasks.

\bibliographystyle{plain}
\bibliography{ref.bib}

\end{document}